\documentclass[]{spie}  %>>> use for US letter paper
\usepackage{amsmath,amsfonts,amssymb}
\usepackage{graphicx}
\usepackage{subcaption}
\usepackage[colorlinks=true, allcolors=blue]{hyperref}

\title{Automatic Microscopic Cell Counting by Use of Deeply-Supervised Density Regression Model}

\author[1]{Shenghua He}
\author[2,3]{Kyaw Thu Minn}
\author[3]{Lilianna Solnica-Krezel}
\author[1,2,4]{Mark Anastasio}
\author[4]{Hua Li}
\affil[1]{Department of Computer Science and Engineering}
\affil[2]{Department of Biomedical Engineering}
\affil[3]{Department of Developmental Biology}
\affil[4]{Department of Radiation Oncology \break Washington University in St$.\,$Louis, St$.\,$Louis, MO, USA}

\authorinfo{Further author information: (Send correspondence to Hua Li \& Mark Anastasio)\\E-mail: \{li.hua, anastasio\}@wustl.edu, Telephone: 1 314 537 7145 \& 1 314 935 3637}

\begin{document} 
\maketitle

\begin{abstract}
Accurately counting cells in microscopic images is important for medical diagnoses and biological studies, but manual cell counting is very tedious, time-consuming, and prone to subjective errors, and automatic counting can be less  accurate than desired. To improve the accuracy of automatic cell counting, we propose here a novel method that employs deeply-supervised density regression. %There remains an important need to improve the accuracy of automated cell counting methods.
%An automatic and efficient solution to improve the cell counting accuracy is critical for any cell counting task. 
%Automatic cell counting is a challenging task due to the low image contrast, strong tissue background, and significant inter-cell occlusions in microscopy images. 
%In this study, a novel density regression-based method is proposed for automatic cell counting.
A fully convolutional neural network (FCNN) serves as the primary FCNN for density map regression.
% model (DRM) for estimating the density map of a given image. 
%In order to improve the ability of the intermediate layers in PriCNN to extract more informative features for accurate cell counting, 
Innovatively, a set of auxiliary FCNNs are employed to provide additional supervision for learning the intermediate layers of the primary CNN to improve network performance. 
In addition, the primary CNN is designed as a concatenating framework to integrate multi-scale features through shortcut connections in the network, 
which improves the granularity of the features extracted from the intermediate CNN layers 
and further supports the final density map estimation.
%which can further improve the performance of the PriCNN for automatic cell counting.
%Both the PriCNN and AuxCNNs are cooperatively optimized.
% through the minimization of the loss functions measured at their outputs. 
%After the training, the trained PriCNN is employed to estimate the density map for a given to-be-test image, and the number of cells can be obtained from the estimated density map. 
The experimental results on immunofluorescent images of human embryonic stem cells demonstrate the superior performance of the proposed method over other state-of-the-art methods.
\end{abstract}


\section{INTRODUCTION}
\label{sec:intro}  % \label{} allows reference to this section

Accurately counting the number of cells in microscopic images greatly aids medical diagnoses and biological studies~\cite{coates2015tailoring}. 
%For example, the number of proliferating tumor cells in breast area help doctors in better understanding the breast tumor and exploring various treatment options~\cite{coates2015tailoring}. 
However, manual cell counting, which is slow, expensive, and prone to subjective errors, is not practically feasible for high-throughput processes. 
%There remains an important need to improve the accuracy of automated cell counting methods.
%An automatic cell counting method is critical for cell counting tasks.
An automatic and efficient solution with improved counting accuracy is highly desirable, but automatic cell counting is challenging due to the low image contrast, strong tissue background, and significant inter-cell occlusions in 2D microscopic images~\cite{matas2004robust,barinova2012,arteta2012,xing2014automatic}.

To address these challenges~\cite{lempitsky2010,xie2018}, density-based counting methods have received increasing attention, due to their superior performance to those traditional detection-based methods~\cite{arteta2016, cirecsan2013, xing2014automatic, liu2017}.
Generally speaking, density-based methods employ machine-learning tools to learn a density regression model (DRM) that estimates the cell density distribution from the characteristics/features of a given image.
The number of cells can be subsequently estimated by integrating the estimated density map.
%Several density regression models (DRMs) have been proposed in the literature~\cite{lempitsky2010,xie2018}.
For example, Lempitsky et al.~\cite{lempitsky2010} proposed a supervised learning framework 
to learn a linear DRM and employ it for visual object counting tasks.
%The learning process is formulated as a minimization of a regularized risk quadratic cost function.
%The number of objects is then calculated as the integral across over the learned density map.
Differently, Xie et al.~\cite{xie2018} utilized a fully convolutional regression network (FCRN) to learn a DRM for regressing cell spatial densities over the image.
The FCRN, as a specific fully convolutional neural network (FCNN), integrates informative feature extraction and powerful function learning to estimate the density map of a given image.
It demonstrates promising performance for cell counting tasks, especially for counting overlapped cells.

Generally, the layers in the FCRN are hierarchically constructed, and the output of each layer relies on the outputs of its previous layers.
There are two potential shortcomings of this traditional network design. 
First, the intermediate layers are optimized based on the gradient information back-propagated only from the final layer of the network, not directly from the adjacent layer.
Second, this design allows only adjacent layers to be connected, while limiting the integration of multi-scale features (or information) and overall network performance.
These two shortcomings might lead to sub-optimized intermediate layers, 
and can eventually affect the overall cell counting accuracy. Thus, the need to improve the accuracy of automated cell counting methods remains.

Recently, in order to improve the effectiveness of learning intermediate layers in a designed deep neural networks, deeply-supervised learning (or deep supervision) has been proposed and has shown promising performance for addressing various computer vision tasks, such as classification~\cite{lee2015deeply} and segmentation~\cite{zeng20173d,dou20173d}.
In addition, concatenating CNN frameworks have also attracted great attention.
These networks can concatenate multi-scale features by shortcut connections of non-adjacent layers within the network, and so achieve better results than the traditional networks in such computer vision tasks as segmentation~\cite{ronneberger2015} and detection~\cite{dong2017automatic}.

Motivated by these works, this study proposes a novel density regression-based automatic cell counting method.
A FCNN is used as a primary FCNN (PriCNN) to learn the density regression model (DRM) that performs an end-to-end mapping from a cell image to the corresponding density map.
%Different from those previous FCNN-based methods which can only indirect supervise the training of intermediate layers based on the feedback from the final network layer, 
A set of auxiliary CNN (AuxCNNs) are built to assess the features at the intermediate layers in the PriCNN and to directly supervise the training of these layers.
In addition, by use of concatenation layers, the multi-scale features from non-adjacent layers are integrated to improve the granularity of the features extracted from the intermediate layers for further supporting final density map estimation.
Experimental results, evaluated on a set of immunofluorescent images of human embryonic stem cells (hESC), 
have demonstrated the superior performance of the proposed deep supervision-based DRM method compared to other state-of-the-art methods.

\section{Methodology}
\label{sec:methodology}

\subsection{Background: Density-Based Automatic Cell Counting}
\label{ssec:density}

The goal of density regression-based cell counting methods is to learn a density regression function $F$, 
which can be employed to estimate the density map of a given image~\cite{lempitsky2010,xie2018}.
Given an image $X\in \mathbb{R}^{M\times N}$ which includes $N_c$ cells,
the density map $Y\in\mathbb{R}^{M\times N}$ of $X$ 
can be considered as the superposition of a set of $N_c$ normalized 2D discrete Gaussian kernels that are placed at the centroids of the $N_c$ cells.
Therefore, the number of cells can be counted by integrating the density map over the image.

%Let $S\in \mathbb{R}^{M\times N}$ represent the cell centroid positions in $X$, then each pixel $Y_{i,j}$ on the density map $Y$ of $X$ can be expressed as:
%%
%\begin{equation}
%Y_{i,j} = \sum_{k=1}^{N_c} \boldsymbol{\mathcal{N}} (s_{k}; m, \sigma^{2}), \quad where \quad i \in M, j \in N,
%\label{eq:map}
%\end{equation}
%%
%\noindent where $\boldsymbol{\mathcal{N}}(s_k; m, \sigma^{2})$ is a normalized discrete 2D Gaussian kernel with a mean $m$ and an isotropic covariance $\sigma^2=(\sigma_0^2,\sigma_0^2)$, and is applied at a centroid location $s_k \in \mathbb{N}^2, k = 1,2,...,N_c$.
Let $S = \{ ({s_{k_x}},{s_{k_y}})\in \mathbb{N}^2: k = 1,2, ..., N_c\}$ represent the cell centroid positions in $X$. Each pixel $Y_{i,j}$ on the density map $Y$ can be expressed as:
\begin{equation}
Y_{i,j} = \sum_{k=1}^{N_c} G_\sigma (i-{s_{k_x}},j-{s_{k_y}}), \quad \forall \quad i \in M, j \in N, \\
\label{eq:map}
\end{equation}
\noindent where $G_\sigma(n_x,n_y) = C \cdot e^{-\frac{n_x^2+n_y^2}{2\sigma^2}}\in \mathbb{R}^{(2K_G+1)\times (2K_G+1)}, n_x, n_y = -K_G, -K_G+1, ..., K_G$, is a normalized 2D Gaussian kernel that satisfies $\sum_{n_x=-K_G}^{K_G}\sum_{n_y=-K_G}^{K_G} G_\sigma(n_x,n_y) =1$. Here, $\sigma^2$ is the isotropic covariance, $(2K_G+1)$ is the kernel size, and $C$ is a normalization constant.

The density regression-based cell counting process generally includes three steps: (1) map an image to a feature map, (2) estimate a cell density map from the feature map, and (3) integrate the density map for automatic cell counting.
In the first step, each pixel $X_{i,j}$ in $X$ can be assumed to be associated with a real-valued feature vector $\phi(X_{i,j})\in R^Z$. 
The feature map $P\in \mathbb{R}^{M\times N \times Z}$ of $X$ can be generated using specific feature extraction methods, such as the dense scale invariant feature transform (SIFT) descriptor~\cite{vedaldi2010vlfeat}, ordinary filter banks~\cite{fiaschi2012}, or codebook learning~\cite{sommer2011ilastik}.
In the second step, the estimated density $\hat{Y}_{i,j}$ of each pixel $X_{i,j}$ in $X$ can be obtained by applying a pre-trained density regression function $F$ on the given $\phi(X_{i,j})$:
\begin{equation}
	\hat{Y}_{i,j} = F(\phi(X_{i,j});\Theta),
	\label{eq:estimatemap}
\end{equation}
\noindent where $\Theta$ is a parameter vector that determines the function $F$.
Finally, in the third step, the number of cells in $X$, $N_c$, can be counted by integrating the estimated densities $\hat{Y}$ over the image region: 
\begin{equation}
	N_c \approx \hat{N_c} = \sum_{i=1}^{M}\sum_{j=1}^{N} \hat{Y}_{i,j}.
	\label{eq:counting}
\end{equation}

A key task in density regression-based cell counting methods is learning the function $F$ by use of training
datasets.
The learning of $F$ and the related cell counting method proposed in this study are described below.

%\textcolor{red}{A set of training images and the associated ground-truth density maps is employed to learn the function $F$ by use of supervised learning.}
%Metrics such as mean square error (MSE)~\cite{lempitsky2010}, mean absolute error (MAE)~\cite{lempitsky2010,fiaschi2012,arteta2014}, and maximum excess over subarrays (MESA)~\cite{lempitsky2010}, have been employed to measure the average error between the estimated density maps and their corresponding ground truths.}

\subsection{The Proposed Automatic Cell Counting Framework}
\label{ssec:density}

\begin{figure}
	\centering
	\includegraphics[width=\textwidth]{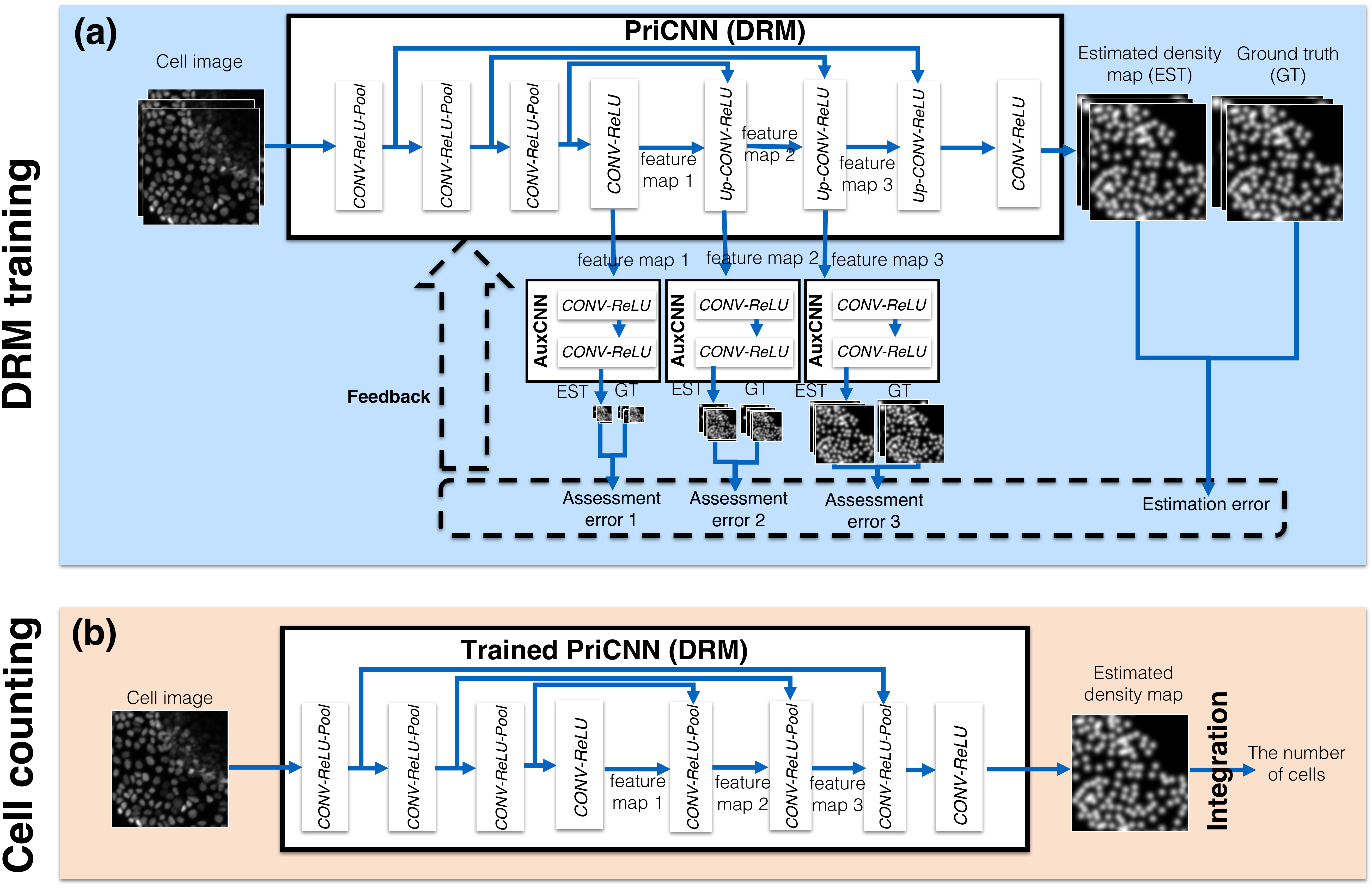}
	\caption{The framework of the proposed automatic cell counting method. The first phase (a)  is DRM training, and the second phase (b) is cell counting by use of the trained DRM.
%		 by : FCNN-based DRM training, and density estimation and cell counting. The FCNN consists of a PriCNN, and a set of AuxCNNs. In the figures, a ReLU layer is placed after every CONV layer but not explicitly presented.
%\textcolor{red}{same here, the figure is so complex, and not clear at all. 1. when you mean DRM, Only the PriCNN is DRM or the combination of PriCNN and AuxCNN is the DRM, need to be very clear, Or AuxCNN is only used to help the training of PriCNN (which is the DRM)?
%Again, arrows pointing out from PriCNN to AuxCNN, but AuxCNN point out the image or density map, whats the relationship between these pointed out image or density map to the method? If having iinput for a block, there should have an output to soemwhere, in this figure 2, you have density map 128x128, 64x64, blahblah, in my opinion, no matter what, you should have only one final output, no one understand why so many output here. the figure needs to be very clear}
		}
	\label{fig:framework}
\end{figure}

We propose a novel automatic cell counting method that employs deeply-supervised density regression model in this study.
The framework, shown in Figure~\ref{fig:framework}, includes two phases: 1) DRM training and 2) cell counting by use of the trained DRM.
The network architecture of the proposed DRM is described in Section~\ref{sssec:drm}. 
The two phases in the framework are described in Sections~\ref{sssec:drm-training} and~\ref{sssec:counting}, respectively.
%		 by : FCNN-based DRM training, and density estimation and cell counting. The FCNN consists of a PriCNN, and a set of AuxCNNs. In the figures, a ReLU layer is placed after every CONV layer but not explicitly presented.
%\textcolor{red}{same here, the figure is so complex, and not clear at all. 1. when you mean DRM, Only the PriCNN is DRM or the combination of PriCNN and AuxCNN is the DRM, need to be very clear, Or AuxCNN is only used to help the training of PriCNN (which is the DRM)?
%Again, arrows pointing out from PriCNN to AuxCNN, but AuxCNN point out the image or density map, whats the relationship between these pointed out image or density map to the method? If having iinput for a block, there should have an output to soemwhere, in this figure 2, you have density map 128x128, 64x64, blahblah, in my opinion, no matter what, you should have only one final output, no one understand why so many output here. the figure needs to be very clear}
%

\subsubsection{The DRM Network Architecture}
\label{sssec:drm}

The DRM is built as a primary FCNN (PriCNN) with the purpose of estimating the density map $\hat{Y}$ 
of an image $X$, such that:
\begin{equation}
Y \approx \hat{Y} = F(X;\Theta),
%Y \approx \hat{Y} = F(X),
\end{equation}
where $F(X;\Theta)$ is a density regression function, and $\Theta$  is a parameter vector that determines $F$.

Motivated by the network architecture of FCRN~\cite{xie2018}, the designed PriCNN (DRM) includes $8$ chained blocks.
As shown in the upper row of Figure~\ref{fig:framework}, 
each of the first three blocks includes a convolutional (CONV) layer, a ReLU layer, and a max-pooling (Pool) layer; 
the fourth block in the PriCNN includes a CONV layer and a ReLU layer; 
each of the fifth to seventh blocks includes a up-sampling (UP) layer, a CONV layer, and a ReLU layer; 
and the last block includes a chain of a CONV layer and a ReLU layer.
%The CONV layer, associated with a learnable kernel, is employed to extract local features from the output of the previous layer.
%\textcolor{red}{
%A ReLU layer sets negative responses to zero and keep only positive responses, and is used to enforce nonlinearity in the network. 
%A Pool layer performs a down-sampling operation and outputs only the maximum value in every down-sampled region, so that the spatial size of the input feature map is progressively reduced but the informative features are left. 
%An Up layer performs an up-sampling operation to restore the resolution of the final estimated density map.}

In addition, concatenation layers are employed in the PriCNN to integrate multi-scale features and thus improve the granularity of the features, which assists in the final density map estimation.
This design is motivated by a network architecture described by Ronneberger et. al~\cite{ronneberger2015}.
As shown in Figure~\ref{fig:framework}, the outputs from each of the first three blocks are multi-resolution, low-dimension, and highly-representative feature maps.
Three shortcut connections are established to connect the first and seventh blocks, the second and sixth blocks, and the third and fifth blocks, respectively. 
With these shortcut connections, multi-resolution features can be concatenated between non-adjacent layers. 
The integration of multi-scale features can further improve the performance of the network 
compared to the traditional FCRN, which allows only adjacent layers to be connected.

%the fully-connected layer that is directly fed to the output layer, which improves the integration of . 
%The efficiency and XXX of the network can be improved.
%Symmetrically, the chain of the last four blocks generate an estimated density map with the same size of the given image.
%In addition, the FCRN employed in the literature\textcolor{red}{REF} only allows adjacent layers connected, 
%which limits the integration of multi-scale information.
%Different to that, concatenation layers are employed in the PriCNN to improve the granularity of the features for the final estimation, motivated by the network architecture employed in the literature~\cite{ronneberger2015}.
%As shown in Figure~\ref{fig:framework}, the first block is connected with both the second and seventh blocks. 
%The similar connections are employed for the second and third blocks  
%In this way, multi-scale features can be concatenated by shortcut connections to the
%fully-connected layer that is directly fed to the output layer. 
%The efficiency and XXX of the network can be improved.

\subsubsection{DRM Training Process}
\label{sssec:drm-training}

\textit{\textbf{1. AuxCNN-supported DRM Training}}

Training the designed DRM (or PriCNN) with such a hierarchical structure is a challenging task. 
%The traditional methods[ref] such as XXX, XXX, and XXX learning all the layers (or block?) based on the feedback from the final layers.
As described in Section~\ref{sec:intro}, all the layers in the original FCRN~\cite{xie2018} are learned based on the feedback only from the final layer. 
Therefore, the intermediate layers might be sub-optimized, which can significantly affect the accuracy of the final estimated density map.

\begin{figure}
	\centering
	\includegraphics[width=\textwidth]{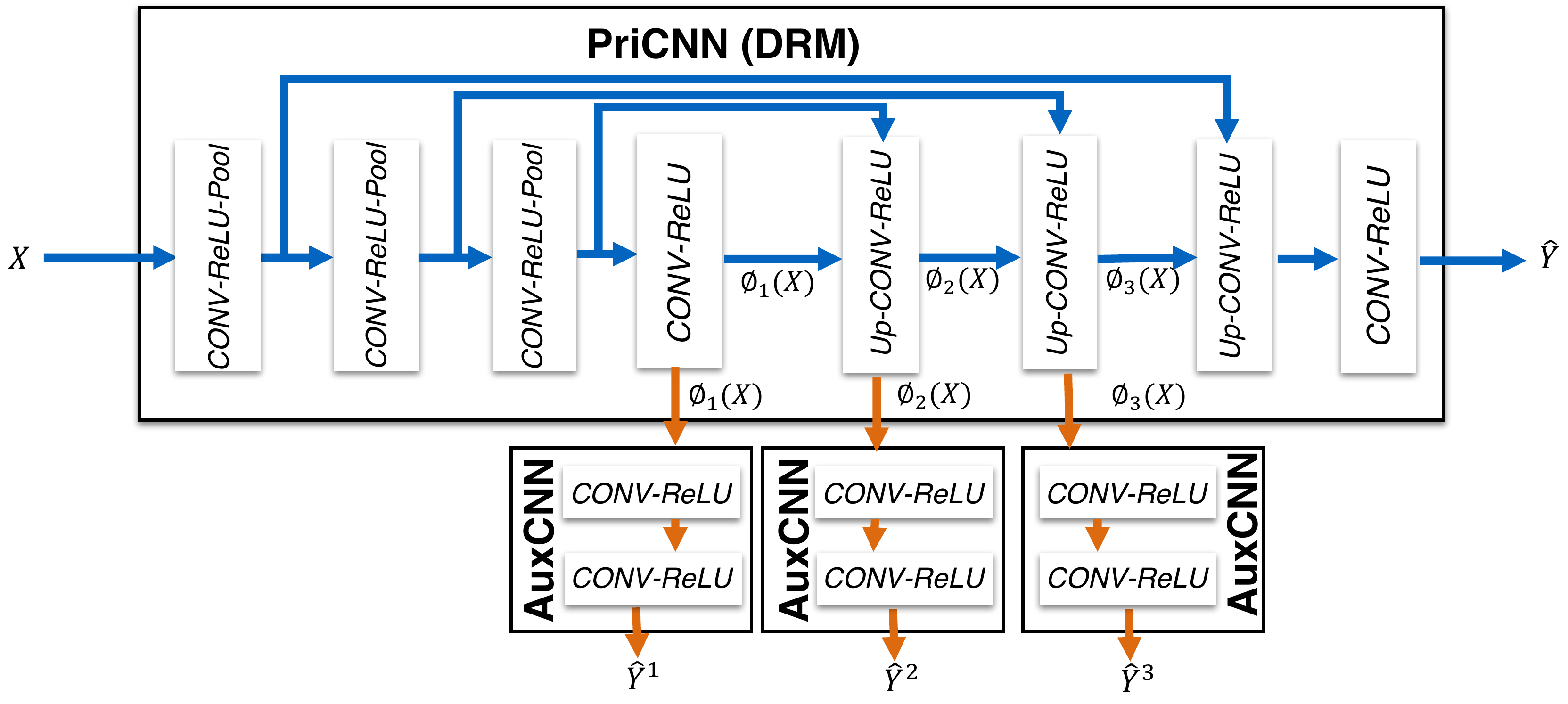}
	\caption{The network architectures of the PriCNN and 3 AuxCNNs in the proposed approach. 
%The green and red dash rectangular frames respectively represent the architecture of the PriCNN and those of the AuxCNNs. Each solid blue box corresponds to a multi-channel feature map. The number of channels is denoted on top of the box. The spatial size is provided at the left edge of the box. White boxes represent copied feature maps, and they are concatenated with the blue boxes as the input of the next layer. The arrows denote the different operations. The kernel size are denoted after CONV operation. A ReLU layer is placed after each CONV layer, and not explicitly mentioned in the figure. The translucent boxes represent the four stages of layers in the PriCNN.
%\textcolor{red}{both this figure and the other figure need to be revised. too complex, not clear at all. Also, see the figure, you have arrows pointing from Pricnn to AnxCNN, then what\rq{}s the output of AuxCNN, to where? help what?}
	}
	\label{fig:architecture}
\end{figure}

%In order to improve the learning quality of the intermediate layers in the PriCNN for accurate cell counting, 
Innovatively, three auxiliary FCNNs (AuxCNNs) are employed to provide additional supervision for learning the intermediate layers of the PriCNN.
As shown in Figure~\ref{fig:architecture}, each AuxCNN contains two CONV-ReLU blocks for estimating 
a low-resolution density map from the feature map generated at an intermediate layer of the PriCNN.
By jointly minimizing the errors between the estimated density maps and the corresponding ground truth density maps at different resolution levels, 
the optimization of the intermediate layers in the PriCNN can be improved, 
which eventually improves the overall performance of the PriCNN.

\noindent {\textit{\textbf{2. Jointly Training the PriCNN and AuxCNN}}}

The parameters of the PriCNN can be learned by jointly training the PriCNN and the AuxCNNs with 
a set of given training data $D = \{(X_i,Y_i)\}_{i = 1,2,..,B}$, 
where $X_{i}$ and $Y_i$ represent the $i$-th image and its associated ground-truth density map, respectively.
The training is completed by minimizing
the differences between the estimated density maps at different resolution levels and the ground truth density maps.

As shown in Figure~\ref{fig:architecture}, the PriCNN can be denoted mathematically as $F(X;\Theta)$, 
where $\Theta$ is the parameter vector of $F$ and $X$ is an input image.
All the trainable parameters in the first four blocks, the $5$th, the $6$th, and the last two blocks can be denoted as 
$\Theta_1$, $\Theta_2$, $\Theta_3$, and $\Theta_4$, respectively. 
Therefore, $\Theta = (\Theta_1, \Theta_2, \Theta_3, \Theta_4)$ and $F(X;\Theta) = F(X;\Theta_1, \Theta_2, \Theta_3, \Theta_4)$.
Also, the output feature maps of the $4$-th, $5$-th, and $6$-th blocks can be denoted as 
$\phi_1(X;\Theta_1)$, $\phi_2(X;\Theta_1,\Theta_2)$, and $\phi_3(X;\Theta_1,\Theta_2,\Theta_3)$, respectively.
Similarly, the three AuxCNNs can be denoted as $A_k(\phi_k;\theta_k) (k = 1,2,3)$, where $\theta_k$ is the parameter vector of the $k$-th AuxCNN, and $A_{k}$ is the density map estimated by use of the $k$-th AuxCNN.
Therefore, the cooperative training of the PriCNN and AuxCNNs is performed by jointly minimizing four loss functions, defined below:
\begin{equation}
\begin{aligned}
\begin{cases}
L_k (\Theta_1, ..., \Theta_k, \theta_k) &= \frac{1}{B}\sum_{i=1}^{B} \left\| A_k (\phi_k(X_i;\Theta_1,...,\Theta_k);\theta_k) - Y^k_i \right\|^2, k = 1,2,3,\\
L(\Theta) &= \frac{1}{B}\sum_{i=1}^{B} \left\| F(X_i,\Theta) -Y_i \right\|^2,
\end{cases}
\end{aligned}
\end{equation}
\noindent where $Y^k_i$ is the ground truth low-resolution density map (GTLR) generated from $Y_i$.
$Y^k_i\in \mathbb{R}^{M_k\times N_k}$ is generated from the original ground-truth density map $Y_i\in \mathbb{R}^{M\times N}$ by summing every adjunct $a\times b$ in $Y$, with $a_k = \frac{M}{M_k}$ and $b_k= \frac{N}{N_k}$.
%Here, $(a_1, b_1), (a_2, b_2)$, and $(a_3,b_3)$ are set to $(8,8)$, $(4,4)$, and $(2,2)$ respectively. 
$L(\Theta)$ is the average mean square error (MSE) between the estimated density maps and their ground truths.
$L_k (\Theta_1, ..., \Theta_k, \theta_k)$ is the average MSE between the low-resolution density maps estimated by $k$-th AuxCNN and their corresponding GTLR density maps.

To improve the computational efficiency of the optimization of the PriCNN and AuxCNNs,
we construct a combined loss function, defined as below:
\begin{equation}
L_{overall}(\Theta, \theta_1, \theta_2, \theta_3) = L(\Theta)+\sum_{k = 1}^{3} \lambda_k L_k (\Theta_1, ..., \Theta_k, \theta_k),
\label{eq:loss}
\end{equation}
where $\lambda_k \in [0,1]$ is a parameter that controls the relative strength of the supervision 
under the $k$-th AuxCNN for learning the intermediate layers in the PriCNN. 
%+ \lambda (\left\| \Theta \right\|^2 + \sum_{k=1}^{3} \left\| \theta_k \right\|^2),
Eqn.(\ref{eq:loss}) is numerically minimized via stochastic gradient descent (SGD) methods~\cite{bottou2010large}.

\subsubsection{Density estimation and cell counting}
\label{sssec:counting}

During the cell counting phase of the framework (Figure~\ref{fig:framework}), 
the number of cells in a to-be-tested image $X^t\in \mathbb{R}^{M'\times N'}$ can be estimated by use of the trained DRM represented by $F(X;\Theta^*)$:
\begin{equation}
\hat{N_c} = \sum_{i=1}^{M'} \sum_{j=1}^{N'} [F(X^t;\Theta^*)]_{i,j},
\label{eq:test}
\end{equation}

\noindent where $[F(X^t;\Theta^*)]_{i,j}$ is the estimated density at pixel $(i,j)$. 
%\textcolor{red}{we might need to make sure how to say here. we call F as the density regression function, I revised it, double check} 
In this step, the dimensions of the to-be-tested image can be different because arbitrary input image sizes are allowed by the trained PriCNN.

\section{Experimental Results}
\label{sec:experiment}

\subsection{Dataset}

A set of $49$ immunofluorescent images of human embryonic stem cells (hESC) was employed in this study to test the performance of the proposed method. 
Each image was $512\times 512$ pixels, and the $49$ hESC images were manually annotated by identifying the centroid of each cell within each image.
Statistically, the cell number among these images is about $518\pm316$.
%Each of them contains a hESC population with $518\pm316$ cells, and the centroid of each cell was manually annotated as a $1$-pixel dot.
%In order to train the DRM s for the evaluation of cell counting methods, 

For each annotated image in the training dataset, the corresponding ground truth density maps were generated 
by placing a normalized 2D discrete Gaussian kernel with isotropic covariance, $\sigma^2$, at each annotated cell centroid in the image (details shown in Section~\ref{ssec:density}).
The values of $\sigma$ and $2K_G+1$ were set to $3$ pixels and $21$ pixels, respectively.
A pair consisting of an image and a density map was considered as a training sample for training the PriCNN and AuxCNNs.
In this study, 5-fold cross validation was employed to evaluate the cell counting performance.
%In this study, $35$ out of the $52$ images were employed as the training set, and the remaining $17$ were the testing set.
%\textcolor{red}{This sentence should put in the discussion. The images in the dataset exhibit low image contrast, and large variance in the numbers of cells across images, which makes it challenging to automatically count the number of cells.}

\subsection{Method Implementation}
\label{ssec:implement}

We compared the performance of the proposed method (denoted as PriCNN+AuxCNN) with a state-of-the-art method, FCRN~\cite{xie2018}.
The FCRN used the same network architecture as the PriCNN, but without concatenation layers. 
In addition, a PriCNN without AuxCNNs (or a FCRN with concatenation layers) was also compared to illustrate the performance improvement by use of AuxCNNs.

%Let us denote the proposed method as \textit{PriCNN+AuxCNN}. It was evaluated on the dataset introduced above, and compared with the state of the art, FCRN~\cite{xie2018}. 
%FCRN is a FCNN with the same network architecture as the PriCNN, but without concatenation layers. Furthermore, a PriCNN without AuxCNNs is also trained for comparison, which is denoted as \textit{PriCNN-only}.

In the PriCNN, the convolution kernel size in the first $7$ blocks was set to $3\times 3$,
while that in the last block was set to $1\times 1$.
The numbers of kernels in the first to $8$th CONV layers are set to 32, 64, 128, 512, 128, 64, 32, and 1, respectively. The pooling size in each pool layer was set to $2\times 2$, and 
the Up layers performed bi-linear interpolation.

In the first block of the AuxCNN, the kernel size was set to $3\times3$ and the number of kernels was $32$, while the comparable values in the second block were $1\times 1$ and $1$, respectively.
In addition, the ground truth low-resolution density map (GTLR) $Y^k_i\in \mathbb{R}^{M_k\times N_k}$ was generated from the original ground-truth density map $Y_i\in \mathbb{R}^{M\times N}$ by summing local regions with size of $(a_1, b_1)=(8,8)$, $(a_2, b_2)=(4,4)$, and $(a_3,b_3)=(2,2)$, respectively.

All the three methods, PriCNN+AuxCNN, PriCNN-only, and FCRN were trained under the same hyper-parameter configurations, including a learning rate of $0.0001$ and a batch size of 100. 
In addition, all the parameters were orthogonally initialized~\cite{saxe2013exact}.

\subsection{Results}

\begin{figure}
	\centering
	\includegraphics[width = \textwidth]{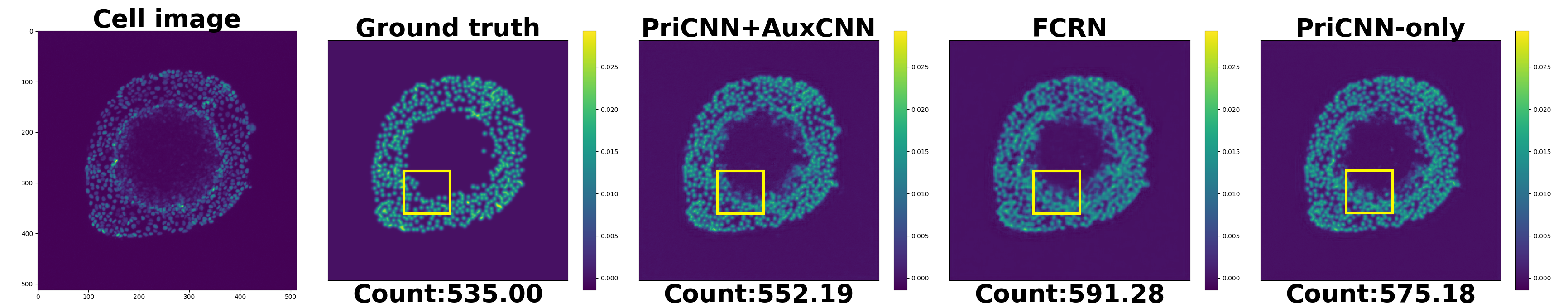}
	\caption{Density estimation for one hESC image example.}
	\label{fig:estimation_samples}
\end{figure}

In this study, mean absolute error (MAE) and standard deviation of absolute errors (STD) were employed to evaluate the cell counting performance. MAE measures the mean of the absolute errors (MAE) between the estimated cell counts and their ground truths for all images in the validation set. The STD measures the standard deviation of the absolute errors.
%The mean absolute error (MAE) and standard deviation (STD) between the estimated numbers of cells and their ground truths on all testing sets is employed to evaluate the performance of the proposed method.
Table~\ref{tab:mae} shows that the proposed method yields superior cell counting performance to the other two methods in terms of MAE and STD.
%The MAE $\pm$ standard deviation (STD) of errors for the evaluated methods are shown in Table~\ref{table:mae_result}. The proposed method shows superior MAE performance, compared with the other two methods.
In addition, Figure~\ref{fig:estimation_samples} presents the estimated density maps of one hESC image example estimated by the three methods. 
The numbers of cells counted from density maps are indicated below each density map. From the figure, we can see that the proposed method (PriCNN+AuxCNN) can estimate a density map that is more similar to the ground truth, compared to the other two methods. Also, our estimated cell count is closer to the ground truth.
%Again, 
%the proposed method achieves better density estimation accuracy.
%\textcolor{red}{Figure 3 font size too small.}
% compared with FCRN, and PriCNN-only.
%
% \setlength{\tabcolsep}{4pt}
%\begin{table}
%\begin{center}
%\caption{MAE$\pm$STD performance for each method.}
%\label{table:mae_result}
%\begin{tabular}{c|ccc}
%\hline\noalign{\smallskip}
%{\bf Methods} & {\bf FCRN~\cite{xie2018}} & {\bf PriCNN-Only} & {\bf PriCNN+AuxCNN}\\
%\noalign{\smallskip}
%\hline
%\noalign{\smallskip}
%{\bf MAE } & $40.5$ &$37.81$& ${\bf 30.79}$\\
%\hline
%\noalign{\smallskip}
%{\bf STD} & $49.35$ &$35.16$& ${\bf 31.31}$\\
%\hline
%\end{tabular}
%\end{center}
%\end{table}
%\setlength{\tabcolsep}{1.4pt}

\begin{table}[ht]
\caption{Performance of the proposed cell counting method} 
\begin{center}       
\begin{tabular}{c c c c} 
\hline\hline
Performance  	& PriCNN+AuxCNN 		& PriCNN-only 			& FCRN~\cite{xie2018}\\
MAE 	&  \textbf{32.89}	& 42.17 	& 44.90\\
STD	&  \textbf{26.35}	       & 30.97 	& 35.30\\
\hline\hline
\end{tabular}
\end{center}
\label{tab:mae}
\end{table}

\section{Discussion}
\label{sec:discussion}
%In this work, a deeply-supervised density regression framework is proposed for automatic microscopic cell counting. A fully convolutional neural network (FCNN) serves as the primary FCNN for density map regression. Innovatively, a set of auxiliary FCNNs are employed to provide additional supervision for learning the intermediate layers of the primary FCNN to improve network performance. In addition, the primary CNN is designed as a concatenating framework to integrate multi-scale features through shortcut connections in the network, 
%which improves the granularity of the features extracted from the intermediate CNN layers 
%and further supports the final density map estimation.
%
%The effectiveness of the additional supervision from the feature assessments conducted by AuxCNNs can be confirmed by the results shown in Table~\ref{tab:mae}, 
%where PriCNN+AuxCNN shows superior MAE performance to the PriCNN-Only. 
%The comprehensive gradients propagated from errors measured at the outputs of PriCNN and AuxCNNs guide the learning of the PriCNN to a better local minimum, improving the cell counting accuracy.

Convolutional neural networks (CNN) have succeeded in computer vision tasks, including image classification~\cite{he2016}, segmentation~\cite{ronneberger2015, he2018}, and object detection~\cite{ren2015}. The success is because that CNNs can integrate informative feature extraction and powerful nonlinear function learning. Furthermore, fully convolutional neural networks (FCNN), such as FCN~\cite{long2015} and U-Net~\cite{ronneberger2015}, allow flexible input image sizes, and have been employed to perform an efficient end-to-end mapping from an image (one domain) to a probability map (another domain). Both the PriCNN (the proposed DRM in the study) and FCRN~\cite{xie2018} are some specific FCNNs, which explain the descent cell counting accuracy they have achieved in this study.

In this study, only a set of experimental immunofluorescent hESC images was employed for DRM training and validation. However, the generalization of the proposed method should not be limited to only the experimental immunofluorescent hESC images. In future, we will evaluate the proposed method on image sets of other modalities. In addition, other competing general object counting methods will also be compared with our proposed method.

%\textcolor{red}{what you want to express in this paragraph?} Metrics such as mean square error (MSE)~\cite{lempitsky2010,fiaschi2012,arteta2014}, mean absolute error (MAE)~\cite{lempitsky2010}, and maximum excess over subarrays (MESA)~\cite{lempitsky2010}, have been employed to measure the average error between the estimated density maps and their corresponding ground truths for learning the density functions. In this study, the loss function is specifically chosen as MSE.

%\textcolor{red}{what you want to express in this paragraph?}The images in the dataset exhibit low image contrast, and large variance in the numbers of cells across images, which makes it challenging to automatically count the number of cells.

%In this work, limited attention is focused on the tuning of network architectures. The network architecture of the PriCNN involved in this work is built on the conventional FCNN architectures, such as FCRN~\cite{xie2018} and U-Net~\cite{ronneberger2015}. 
%------ Limitation 2 -----
%Secondly, the manual labeling errors are not considered in our work. The images were manually labeled by human experts, and the subjective errors may affect not only the model training but also the accuracies of the method evaluation. Thus, more reliable cell labelling should be also considered in the future work.

\section{Conclusion}
\label{sec:conclusion}
%Automatic cell counting methods are urgently demanded in biomedical field. However, it is a challenging task due to low image contrast, cell clustering, and inter-cell occlusions. 
In this study, for the first time, a deeply-supervised density regression framework is proposed for automatic cell counting. 
The results obtained on experimental hESC images demonstrate the superior cell counting performance of the proposed method, compared with the state of the art.

\acknowledgments % equivalent to \section*{ACKNOWLEDGMENTS}       
This work was supported in part by award NIH R01EB020604, R01EB023045, R01NS102213, and R21CA223799.
 
% References
\bibliography{references} % bibliography data in report.bib

\begin{thebibliography}{10}

\bibitem{coates2015tailoring}
Coates, A.~S., Winer, E.~P., Goldhirsch, A., Gelber, R.~D., Gnant, M.,
  Piccart-Gebhart, M., Th{\"u}rlimann, B., Senn, H.-J., Members, P., Andr{\'e},
  F., et~al., ``Tailoring therapies—improving the management of early breast
  cancer: St gallen international expert consensus on the primary therapy of
  early breast cancer 2015,'' {\em Annals of oncology}~{\bf 26}(8),  1533--1546
  (2015).

\bibitem{matas2004robust}
Matas, J., Chum, O., Urban, M., and Pajdla, T., ``Robust wide-baseline stereo
  from maximally stable extremal regions,'' {\em Image and vision
  computing}~{\bf 22}(10),  761--767 (2004).

\bibitem{barinova2012}
Barinova, O., Lempitsky, V., and Kholi, P., ``On detection of multiple object
  instances using hough transforms,'' {\em IEEE Transactions on Pattern
  Analysis and Machine Intelligence}~{\bf 34}(9),  1773--1784 (2012).

\bibitem{arteta2012}
Arteta, C., Lempitsky, V., Noble, J.~A., and Zisserman, A., ``Learning to
  detect cells using non-overlapping extremal regions,'' in [{\em International
  Conference on Medical Image Computing and Computer-Assisted
  Intervention}{\nolinebreak\hspace{0.1em}]},   348--356, Springer (2012).

\bibitem{xing2014automatic}
Xing, F., Su, H., Neltner, J., and Yang, L., ``Automatic ki-67 counting using
  robust cell detection and online dictionary learning,'' {\em IEEE
  Transactions on Biomedical Engineering}~{\bf 61}(3),  859--870 (2014).

\bibitem{lempitsky2010}
Lempitsky, V. and Zisserman, A., ``Learning to count objects in images,'' in
  [{\em Advances in neural information processing
  systems}{\nolinebreak\hspace{0.1em}]},   1324--1332 (2010).

\bibitem{xie2018}
Xie, W., Noble, J.~A., and Zisserman, A., ``Microscopy cell counting and
  detection with fully convolutional regression networks,'' {\em Computer
  methods in biomechanics and biomedical engineering: Imaging \&
  Visualization}~{\bf 6}(3),  283--292 (2018).

\bibitem{arteta2016}
Arteta, C., Lempitsky, V., Noble, J.~A., and Zisserman, A., ``Detecting
  overlapping instances in microscopy images using extremal region trees,''
  {\em Medical image analysis}~{\bf 27},  3--16 (2016).

\bibitem{cirecsan2013}
Cire{\c{s}}an, D.~C., Giusti, A., Gambardella, L.~M., and Schmidhuber, J.,
  ``Mitosis detection in breast cancer histology images with deep neural
  networks,'' in [{\em International Conference on Medical Image Computing and
  Computer-assisted Intervention}{\nolinebreak\hspace{0.1em}]},   411--418,
  Springer (2013).

\bibitem{liu2017}
Liu, F. and Yang, L., ``A novel cell detection method using deep convolutional
  neural network and maximum-weight independent set,'' in [{\em Deep Learning
  and Convolutional Neural Networks for Medical Image
  Computing}{\nolinebreak\hspace{0.1em}]},   63--72, Springer (2017).

\bibitem{lee2015deeply}
Lee, C.-Y., Xie, S., Gallagher, P., Zhang, Z., and Tu, Z., ``Deeply-supervised
  nets,'' in [{\em Artificial Intelligence and
  Statistics}{\nolinebreak\hspace{0.1em}]},   562--570 (2015).

\bibitem{zeng20173d}
Zeng, G., Yang, X., Li, J., Yu, L., Heng, P.-A., and Zheng, G., ``3d u-net with
  multi-level deep supervision: fully automatic segmentation of proximal femur
  in 3d mr images,'' in [{\em International Workshop on Machine Learning in
  Medical Imaging}{\nolinebreak\hspace{0.1em}]},   274--282, Springer (2017).

\bibitem{dou20173d}
Dou, Q., Yu, L., Chen, H., Jin, Y., Yang, X., Qin, J., and Heng, P.-A., ``3d
  deeply supervised network for automated segmentation of volumetric medical
  images,'' {\em Medical image analysis}~{\bf 41},  40--54 (2017).

\bibitem{ronneberger2015}
Ronneberger, O., Fischer, P., and Brox, T., ``U-net: Convolutional networks for
  biomedical image segmentation,'' in [{\em International Conference on Medical
  image computing and computer-assisted
  intervention}{\nolinebreak\hspace{0.1em}]},   234--241, Springer (2015).

\bibitem{dong2017automatic}
Dong, H., Yang, G., Liu, F., Mo, Y., and Guo, Y., ``Automatic brain tumor
  detection and segmentation using u-net based fully convolutional networks,''
  in [{\em Annual Conference on Medical Image Understanding and
  Analysis}{\nolinebreak\hspace{0.1em}]},   506--517, Springer (2017).

\bibitem{vedaldi2010vlfeat}
Vedaldi, A. and Fulkerson, B., ``Vlfeat: An open and portable library of
  computer vision algorithms,'' in [{\em Proceedings of the 18th ACM
  international conference on Multimedia}{\nolinebreak\hspace{0.1em}]},
  1469--1472, ACM (2010).

\bibitem{fiaschi2012}
Fiaschi, L., K{\"o}the, U., Nair, R., and Hamprecht, F.~A., ``Learning to count
  with regression forest and structured labels,'' in [{\em Pattern Recognition
  (ICPR), 2012 21st International Conference on}{\nolinebreak\hspace{0.1em}]},
   2685--2688, IEEE (2012).

\bibitem{sommer2011ilastik}
Sommer, C., Straehle, C.~N., Koethe, U., Hamprecht, F.~A., et~al., ``Ilastik:
  Interactive learning and segmentation toolkit.,'' in [{\em
  ISBI}{\nolinebreak\hspace{0.1em}]},   {\bf 2}(5),  8 (2011).

\bibitem{bottou2010large}
Bottou, L., ``Large-scale machine learning with stochastic gradient descent,''
  in [{\em Proceedings of COMPSTAT'2010}{\nolinebreak\hspace{0.1em}]},
  177--186, Springer (2010).

\bibitem{saxe2013exact}
Saxe, A.~M., McClelland, J.~L., and Ganguli, S., ``Exact solutions to the
  nonlinear dynamics of learning in deep linear neural networks,'' {\em arXiv
  preprint arXiv:1312.6120}  (2013).

\bibitem{he2016}
He, K., Zhang, X., Ren, S., and Sun, J., ``Deep residual learning for image
  recognition,'' in [{\em Proceedings of the IEEE conference on computer vision
  and pattern recognition}{\nolinebreak\hspace{0.1em}]},   770--778 (2016).

\bibitem{he2018}
He, S., Zheng, J., Maehara, A., Mintz, G., Tang, D., Anastasio, M., and Li, H.,
  ``Convolutional neural network based automatic plaque characterization for
  intracoronary optical coherence tomography images,'' in [{\em Medical Imaging
  2018: Image Processing}{\nolinebreak\hspace{0.1em}]},   {\bf 10574},
  1057432, International Society for Optics and Photonics (2018).

\bibitem{ren2015}
Ren, S., He, K., Girshick, R., and Sun, J., ``Faster r-cnn: Towards real-time
  object detection with region proposal networks,'' in [{\em Advances in neural
  information processing systems}{\nolinebreak\hspace{0.1em}]},   91--99
  (2015).

\bibitem{long2015}
Long, J., Shelhamer, E., and Darrell, T., ``Fully convolutional networks for
  semantic segmentation,'' in [{\em Proceedings of the IEEE conference on
  computer vision and pattern recognition}{\nolinebreak\hspace{0.1em}]},
  3431--3440 (2015).

\end{thebibliography}
\bibliographystyle{spiebib} % makes bibtex use spiebib.bst

\end{document}